\documentclass[letterpaper, 10 pt, journal, twoside]{ieeeconf}  % Comment this line out if you need a4paper

\IEEEoverridecommandlockouts              % This command is only needed if 
\usepackage{graphicx} % for pdf, bitmapped graphics files
\usepackage{amsmath} % assumes amsmath package installed
\usepackage{amssymb}  % assumes amsmath package installed
\usepackage{lipsum}
\usepackage{caption}
\usepackage{subcaption}
\usepackage{cite}
\newcommand{\etal}{\textit{et al}. }
\usepackage{bm}
\usepackage{multirow}
\usepackage{hyperref}
\usepackage{xcolor}
\definecolor{green}{HTML}{00B000}

\usepackage[final]{changes}
\setcommentmarkup{{\IfIsColored{\color{authorcolor}}{}[#1]}}
\definechangesauthor[name={RevA}, color=red]{RevA}
\definechangesauthor[name={RevB}, color=blue]{RevB}
\definechangesauthor[name={RevAandB}, color=green]{RevAB}

\newcommand*{\B}[1]{\ifmmode\bm{#1}\else\textbf{#1}\fi}
\DeclareMathOperator*{\argmax}{arg\,max}
\DeclareMathOperator*{\argmin}{arg\,min}
\makeatletter
\newcommand{\subalign}[1]{%
  \vcenter{%
    \Let@ \restore@math@cr \default@tag
    \baselineskip\fontdimen10 \scriptfont\tw@
    \advance\baselineskip\fontdimen12 \scriptfont\tw@
    \lineskip\thr@@\fontdimen8 \scriptfont\thr@@
    \lineskiplimit\lineskip
    \ialign{\hfil$\m@th\scriptstyle##$&$\m@th\scriptstyle{}##$\hfil\crcr
      #1\crcr
    }%
  }%
}
\makeatother

\title{\LARGE \bf
Multimodal Active Measurement for Human Mesh Recovery\\in Close Proximity
}

\author{Takahiro Maeda$^{1*}$, Keisuke Takeshita$^{2}$, Norimichi Ukita$^{1}$, and Kazuhito Tanaka$^{2}$% <-this %
\thanks{Manuscript received: April 9th, 2024; Revised:
July 11th, 2024; Accepted: September 4th, 2024.}
\thanks{This paper was recommended for publication by
Editor Pascal Vasseur upon evaluation of the Associate Editor and Reviewers’
comments.}
\thanks{$^*$Work done during the internship at Frontier Research Center, Toyota Motor Corporation {\tt\small sd21601@toyota-ti.ac.jp}}
\thanks{$^{1}$Toyota Technological Institute, Nagoya, Aichi, Japan.}%
\thanks{$^{2}$R-Frontier Division, Frontier Research Center, Toyota Motor
Corporation, Toyota, Aichi, Japan.}
\thanks{Digital Object Identifier (DOI): see top of this page.}
}

\begin{document}

\maketitle
\markboth{IEEE Robotics and Automation Letters. Preprint Version. Accepted September, 2024}
{Maeda \MakeLowercase{\textit{et al.}}: Human Mesh Recovery in Close Proximity}

%%%%%%%%%%%%%%%%%%%%%%%%%%%%%%%%%%%%%%%%%%%%%%%%%%%%%%%%%%%%%%%%%%%%%%%%%%%%%%%%
\begin{abstract}
For physical human-robot interactions (pHRI), a robot needs to estimate the accurate body pose of a target person. However, in these pHRI scenarios, the robot cannot fully observe the target person's body with equipped cameras because the target person must be close to the robot for physical interaction. This close distance leads to severe truncation and occlusions and thus results in poor accuracy of human pose estimation. For better accuracy in this challenging environment, we propose an \textit{active measurement} and \textit{sensor fusion} framework of the equipped cameras with touch and ranging sensors such as 2D LiDAR. Touch and ranging sensor measurements are sparse but reliable and informative cues for localizing human body parts. In our \textit{active measurement} process, camera viewpoints and sensor placements are dynamically optimized to measure body parts with higher estimation uncertainty, which is closely related to truncation or occlusion. In our \textit{sensor fusion process}, assuming that the measurements of touch and ranging sensors are more reliable than the camera-based estimations, we fuse the sensor measurements to the camera-based estimated pose by aligning the estimated pose towards the measured points. Our proposed method outperformed previous methods on the standard occlusion benchmark with simulated active measurement. Furthermore, our method reliably estimated human poses using a real robot, even with practical constraints such as occlusion by blankets.
\end{abstract}

\begin{figure}[t]
\centering
    \includegraphics[width=0.95\linewidth]{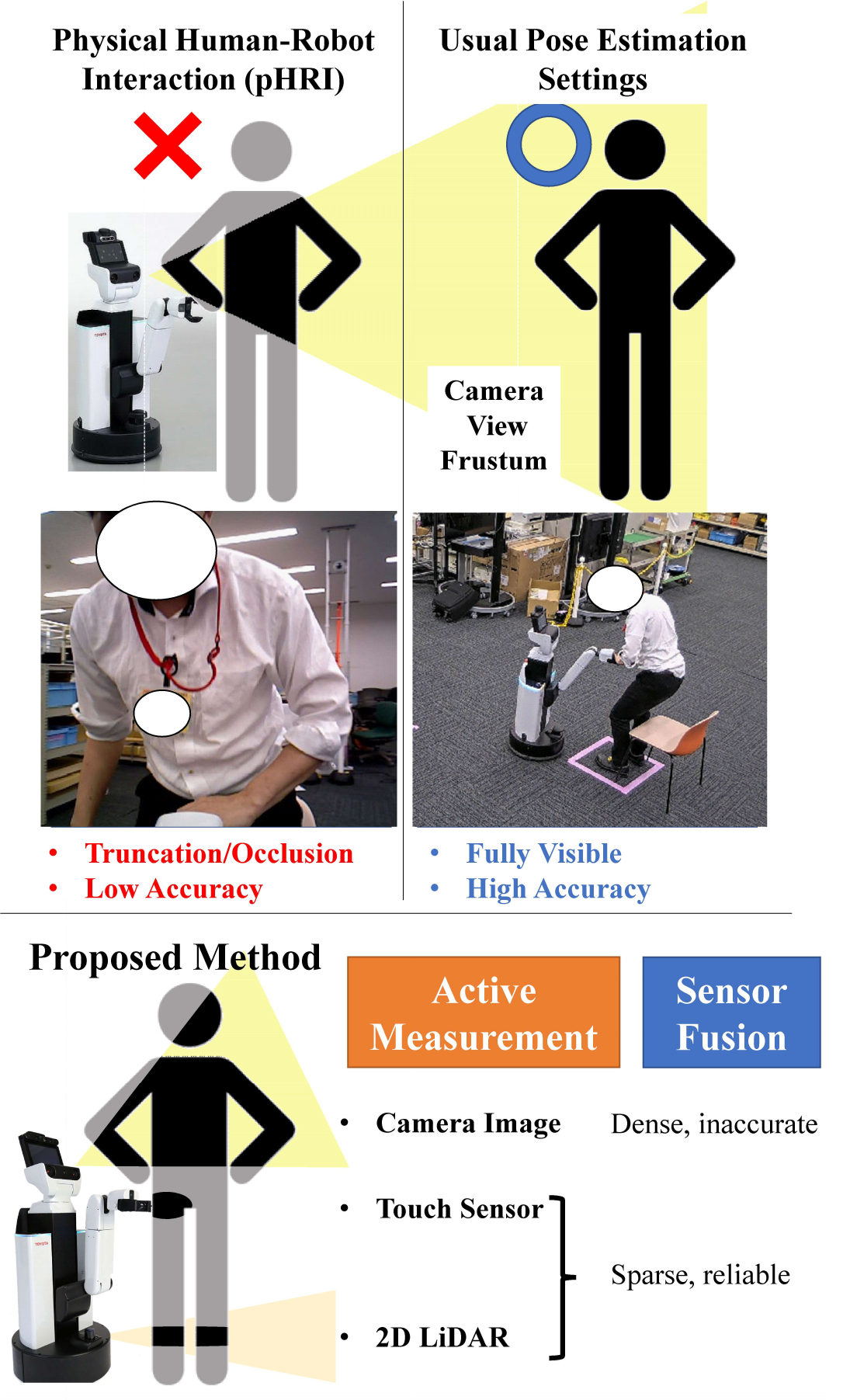}
    \caption{\textbf{Problem of previous human pose estimation methods in pHRI scenarios and our proposed method.}}
\label{fig:teaser}
\end{figure}

%%%%%%%%%%%%%%%%%%%%%%%%%%%%%%%%%%%%%%%%%%%%%%%%%%%%%%%%%%%%%%%%%%%%%%%%%%%%%%%%
\section{INTRODUCTION}

% We excluded the Human-Robot collaboration scenario because HRC mainly focuses on manufacturing settings and our assumption of no additional external cameras doesn't hold

% human pose (or mesh) is necessary for home assistive, especially elderly care and the care of physically challenged people. Or other subjects?

% Possible other pose estimation methods should be discussed in the DISCUSSION section

% Is pose estimation truly difficult in proximal settings? (no depths, no lidar, only partial shot of RGB camera)

% Tactile sensors and force sensors still can be utilized

3D human pose estimation is crucial for assistive robots to physically interact with humans.
For example, to achieve care robots help elderly or physically challenged people stand up stably, robots need the information of the human body to detect unstable postures and falling to decide where to support with their arms.
However, in these physical human-robot interactions (pHRI) scenarios, robots usually cannot fully observe the human body with equipped cameras because occlusions and image truncations happen at a very close range.
Human pose estimation methods perform poorly with these settings due to limited information, as shown in the upper part of Fig.~\ref{fig:teaser}. 

However, humans can still estimate another person's pose in very close range by actively moving their heads and eyes to a more informative viewpoint and using multimodal cues such as tactile information.
% and temperature information from their skin.
Physically interactive robots should perform the same by actively operating their joints and wheels/legs and fuzing multimodal sensor inputs, as shown in the bottom part of Fig.~\ref{fig:teaser}.

% mention that we cannot use depth sensors
% mention that we maximize the coverage during viewpoint selection.
% add agenda

To achieve this, we propose a new human pose estimation method for physically interactive robots.
Our proposed method has two components: \textit{active measurement} and \textit{sensor fusion}.
In the \textit{active measurement}, the camera viewpoints and sensor placements are dynamically optimized to reduce the unobserved body parts by truncations and occlusions. 
%We use the positional variance of estimated poses as the estimation uncertainty, which is closely related to the unobserved body parts due to truncations and occlusions.
%The camera viewpoints are optimized to cover the uncertainty-weighted largest body surfaces.
%Other sensor placements are optimized to measure the most uncertain area within the feasible robot poses.
In the \textit{sensor fusion}, estimated poses from cameras are improved by aligning the body poses consistent with sensor measurements.
%using an optimization algorithm.
Specifically, we minimize the distance between the estimated body surface and the sensor measurement points.
Our contributions are as follows:
\begin{itemize}
    \item We propose the active measurement framework for human pose estimation using multimodal sensors. This framework maximizes the observation area of sensors to cover the whole body.
    \item Our sensor fusion method improves the estimated human pose from optical images by reconfiguring them to coincide with measurements of the human body surface measured from the multimodal sensors.
    \item Our proposed method outperforms previous methods on the standard occlusion benchmark with simulated active measurement.
    We further demonstrate the accuracy improvement of our proposed method with real-world experiments using a robot in application scenarios.
\end{itemize}

\section{RELATED WORK}
% need to add more citations
There has been significant progress in human pose estimation from RGB images, starting from 2D~\cite{DBLP:conf/cvpr/WeiRKS16, DBLP:conf/cvpr/CaoSWS17} to 3D~\cite{DBLP:conf/3dim/MehtaRCFSXT17, DBLP:conf/cvpr/WandtR19, DBLP:conf/nips/ZhangNF20} and mesh recovery~\cite{DBLP:conf/eccv/BogoKLG0B16, DBLP:conf/cvpr/KanazawaBJM18, DBLP:conf/cvpr/0007IMKK22}.
However, most existing methods assume input images are taken from enough distance without occlusions and view-dependent truncations.
Therefore, the following three techniques should be considered in physical human-robot interaction scenarios to deal with occlusions and truncations in close proximity.

\noindent \textbf{Image-based Pose estimation robust to occlusions and view-dependent truncations.}
Some methods utilize attention mechanisms to focus on visible body parts to ensure the robustness to occlusions~\cite{DBLP:conf/icpr/GuWH20, DBLP:conf/iccv/KocabasHHB21, DBLP:conf/cvpr/KhirodkarTK22}.
However, it is difficult to adequately train attention mechanisms due to the scarcity of occluded data.
Data augmentation~\cite{DBLP:journals/pami/XuWLLXZ22, DBLP:conf/cvpr/0005TYFM18} is an easy way for simulating these occluded or truncated data. Still, it cannot fully capture the complexity of real-world occlusions and truncations.
%Besides the previous methods on annotated pose-image pair, we can create pose prior~\cite{DBLP:conf/iccv/ChenSWLY17} as Generative Adversarial Networks~\cite{DBLP:conf/nips/GoodfellowPMXWOCB14} trained on large pose datasets such as AMASS~\cite{DBLP:conf/iccv/MahmoodGTPB19}.
%It can ensure that the output pose is biologically plausible by discriminating poses outside the distribution of the AMASS dataset.
%The early works mainly proposed regression models
%However, it is inherently inappropriate to choose the deterministic regression models for highly ill-posed problems, such as human pose estimation on severely truncated and occluded images.
Some methods use probabilistic models to regress multiple pose hypotheses~\cite{DBLP:conf/iccvw/JahangiriY17, DBLP:conf/nips/BiggsNEJGV20, DBLP:conf/iccv/KolotourosPJD21} for ill-posed problems, such as human pose estimation on severely truncated and occluded images.

Although these methods above can improve estimated poses on ambiguous images, their estimations are just guessing truncated or occluded body parts.
Robots can move to viewpoints with fewer occlusions and truncations in physical human-robot interaction scenarios.

\noindent \textbf{Pose estimation with active measurements.}
The quality of viewpoints is often expressed as estimated error metrics or uncertainty of human pose estimation~\cite{DBLP:conf/icra/ChanKL14, DBLP:conf/cvpr/KicirogluRSSF20}.
Specifically, Kiciroglu \etal~\cite{DBLP:conf/cvpr/KicirogluRSSF20} performs quadratic approximation to error landscape on each viewpoint and treats the approximated variance as uncertainty.
However, this method assumes no truncations and thus fails to model the complex error landscape with severe truncations.
On the other hand, Arzati \etal~\cite{arzati2021viewpoint} optimizes flight drone viewpoints by deep reinforcement learning for dermatology applications.
DQN~\cite{DBLP:journals/corr/MnihKSGAWR13} is trained in simulation environments based on the reward related to pose estimation accuracy.
However, this reward considers only visible body parts and ignores the view-dependent truncated body parts.

Although camera active measurements above can estimate the best camera viewpoint, robots usually have multimodal sensors other than cameras.
Multimodality may improve pose estimation accuracy by complementing limited information from camera observations.
Our proposed active measurement can dynamically optimize camera viewpoints, touch, and ranging sensor positions.

\noindent \textbf{Pose estimation with multimodal data.}
Wearable Inertia Measurement Units (IMUs) attached to human body parts can track their positions and orientations and thus help human pose estimation~\cite{DBLP:journals/cgf/MarcardRBP17, DBLP:journals/tog/HuangKABHP18}.
However, it is inconvenient to wear IMUs for robots' assistance.
Furthermore, wearing IMUs for a long time causes drifting and results in less accurate measurements.

RGB-D cameras and 3D point cloud sensors are often utilized other than RGB cameras~\cite{DBLP:conf/iccv/XiongZ0CYZY19, DBLP:conf/iccv/JiangCZ19, DBLP:conf/cvpr/WangX0LY20}.
Although these sensors can alleviate depth ambiguity and provide accurate estimations, they have a minimum depth distance of around 50 centimeters for measurements, which is inappropriate for human pose estimation in close range.
%Zhao \etal~\cite{DBLP:conf/sigcomm/ZhaoTZALHKK018, DBLP:conf/iccv/ZhaoLRZL0K19} use the unique modality, radio frequency signals, to estimate human poses.
%Although we can estimate the pose of people behind a wall thanks to the characteristics of radio frequency signals, their spatial resolution is low.
% Clever \etal~\cite{DBLP:conf/cvpr/CleverEKTLK20} tried to solve the heavy occlusion on a sleeping person by bed comforters with a pressure sensor mat.

The sensors above are difficult to utilize in pHRI settings due to the amount of effort required to wear IMUs and the minimum depth distances of depth sensors. 
Therefore, we focus on the cues obtained during pHRI, such as touch sensors and 2D LiDARs, which have a minimum distance of around one centimeter, unlike 3D point cloud sensors.
Touch and 2d ranging sensors are sparse compared to 3D point cloud sensors, but reliable cues in pHRI scenarios.
The other possible options are discussed in Sec.~\ref{sec:discussion}.

\section{\added[id=RevA, comment={A.5}]{Preliminary}}

\subsection{SMPL Model}
We represent human poses using Skinned Multi-person Linear Body (SMPL~\cite{DBLP:journals/tog/LoperM0PB15}).
SMPL is a neural network representation $ \mathcal{M}(\bm{\theta}, \bm{\beta}) = V_\text{local}$ that maps a set of pose parameters $\bm{\theta}$ and body parameters $\bm{\beta}$ to a human body mesh in the pelvis-centered local coordinates $V_\text{local} \in \mathbb{R}^{3 \times 6890}$ with 6890 vertices.
The pose parameters $\bm{\theta}$ contain each body joint rotation.
The shape parameters $\bm{\beta}$ control the body height, weights, and limb length.
We can obtain the human body mesh on the world coordinates $V$ using a global pose as $V = R_pV_\text{local} + \bm{t}_p$, where the global pose is a set of global translation and rotation of the pose $(R_p, \bm{t}_p)$.
Global body joint positions $J$ are also regressed by a linear regressor $W$ as $J = R_pWV_\text{local} + \bm{t}_p$.
%We can induce the biological constraints by explicitly using these body models as an intermediate representation.
We use the mesh vertices $V$ to fuse the touch and ranging sensor measurements into the estimated pose, as explained in Sec.~\ref{subsec:fusion}.

\subsection{Probabilistic Human Mesh Recovery}
\label{subsec:hmr}
Human mesh recovery is a task to estimate the set of pose parameters $\bm{\theta}$, body parameters $\bm{\beta}$, and global pose $(R_p, \bm{t}_p)$, which defines human mesh vertices $V$ and eventual joint positions $J$ in the world coordinates, given the camera image $I$ containing a person.
Image-conditioned probability distribution $p(\bm{\theta}, \bm{\beta}|I)$ is obtained by the probabilistic human mesh recovery.
Given this distribution, we can also \textit{regress} the pose via maximum likelihood as follows.
\begin{align}
    &V_\text{ML} = R_p \mathcal{M}(\bm{\theta}_\text{ML}, \bm{\beta}_\text{ML}) + \bm{t}_p \\
    &J_\text{ML} = R_p W \mathcal{M}(\bm{\theta}_\text{ML}, \bm{\beta}_\text{ML}) + \bm{t}_p \\
    &\text{where, } \bm{\theta}_\text{ML}, \bm{\beta}_\text{ML} = \argmax_{\bm{\theta}, \bm{\beta}} p(\bm{\theta}, \bm{\beta}|I)
\end{align}

\section{PROPOSED METHOD}

\begin{figure*}[t]
\centering
\includegraphics[width=\linewidth]{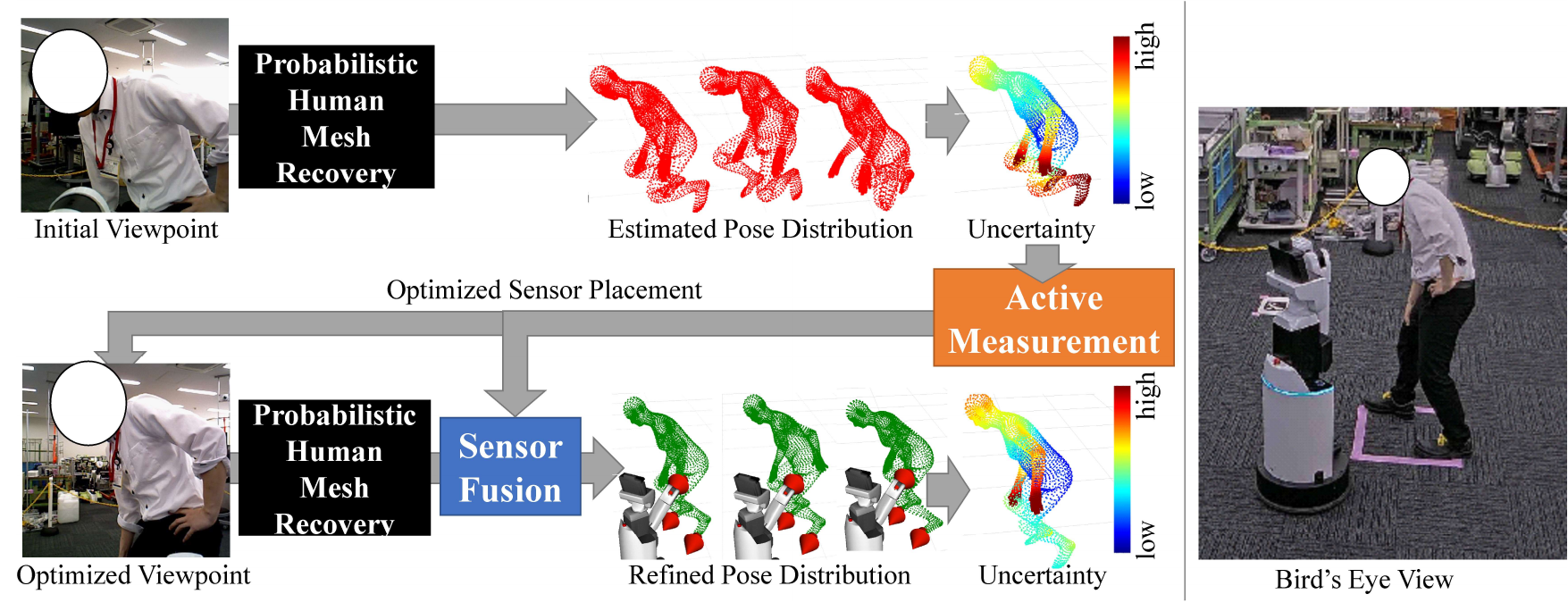}
\caption{\textbf{Overview of our proposed method}.}
\label{fig:proposed_method}
\end{figure*}

The overview of our proposed method is shown in Fig.~\ref{fig:proposed_method}.
Our proposed method is robot-agnostic.

\subsection{Notation}
The objective is to estimate human 3D joint positions $J \in \mathbb{R} ^ {3 \times N_{\text{joint}}}$, where $N_{\text{joint}}$ is the number of joints.
The available information is multimodal measurements including a camera image $I(R_c, \bm{t}_c)$  and touch and ranging $K$ sensor placements $(R_1, \bm{t}_1), \dots, (R_K, \bm{t}_K)$ such as 2D LiDAR and a touch sensor measuring the person's body surface, where $R_c$ and $\bm{t}_c$ are the camera orientation and translation, and $R_k$ and $\bm{t}_k$ are the $k$-th sensor orientation and translation accordingly.
%All sensor positions, including a camera, are optimized by the active measurement framework for better accuracy of human pose estimation.

\subsection{Uncertainty Estimation}
\label{subsec:uncertainty}
We use a probabilistic approach~\cite{DBLP:conf/iccv/KolotourosPJD21} to utilize the estimation uncertainty $\bm{\sigma}^\text{mesh}$ in our active measurement.
The uncertainty of the estimated pose is calculated as positional variances $\bm{\sigma}^\text{mesh} \in \mathbb{R}^{6890}$ of each mesh vertex as follows,
\begin{align}
    \bm{\sigma}^\text{mesh} &= \mathbb{E}_{p(\bm{\theta}, \bm{\beta}|I)} [(\mathcal{M}(\bm{\theta}, \bm{\beta}) - \Bar{\mathcal{M}}(\bm{\theta}, \bm{\beta})^2] \\
    \Bar{\mathcal{M}}(\bm{\theta}, \bm{\beta}) &= \mathbb{E}_{p(\bm{\theta}, \bm{\beta}|I)} [\mathcal{M}(\bm{\theta}, \bm{\beta})]
\end{align}
\added[id=RevB, comment={B.2}]{, where $\mathbb{E}_{p(\bm{\theta}, \bm{\beta}|I)}$ denotes the expectation over the pose distribution conditioned by a image.}
\added[id=RevAB, comment={A.2, B.1}]{There are many examples that utilize predictive variances as uncertainty. One example is a large family of Bayesian Neural Networks~\cite{gal2016dropout, kendall2017uncertainties}. They estimate the aleatoric uncertainty of trained models as predictive variances. This uncertainty of estimated pose $\bm{\sigma}^\text{mesh}$ can account for the occluded or truncated body parts by learning possible pose distributions of observed occluded human images, as shown in Fig.~\ref{fig:proposed_method}. Sengupta~\etal~\cite{sengupta2021hierarchical} showed the per-vertex uncertainty closely correlated to the occluded and truncated body parts.
We chose implicit uncertainty estimation as predictive variances for two reasons. First, predictive variances account for body kinematics using the joint model. For example, the camera-truncated lower leg position is stably predicted if its knee joint is visible. Thus, the uncertainty of the lower leg is low using predictive variances, whereas one can be wrongly high using just explicit camera information. Second, camera view information still needs object detection to estimate the uncertainty from occlusion, whereas predictive variances do not.}
We can determine the effective camera and sensor positions based on the estimated uncertainty $\bm{\sigma}^\text{mesh}$ via our active measurement framework.

\subsection{Active Measurement}
\label{subsec:active_measurement}
The camera viewpoint and sensor placements $(R_c, \bm{t}_c), (R_1, \bm{t}_1), \dots, (R_K, \bm{t}_K)$ are optimized based on the uncertainty of the estimated pose $\bm{\sigma}^\text{mesh}$.
We optimize each placement one by one.
Specifically, after optimizing one sensor, we update the estimated pose and uncertainty $\bm{\sigma}^\text{mesh}$ based on the sensor fusion as described in Sec.~\ref{subsec:fusion}.
We first optimize the camera viewpoint since images usually have the richest information about poses. 

\noindent\textbf{Camera viewpoint optimization:} 
It is difficult to directly optimize camera viewpoints to maximize pose estimation accuracy.
Therefore, we propose to maximize the information obtained from cameras.
Specifically, Camera viewpoint $(R_c, \bm{t}_c)$ is optimized to cover the largest body surface within the camera image and reduce truncation and self-occlusion.
\added[id=RevB, comment={B.3}]{We use weighted coverage with the estimated uncertainty $\bm{\sigma}^\text{mesh}$ to cover possibly more erroneous body parts, which are considered to have higher uncertainty.}
Camera viewpoints are evaluated based on weighted mesh coverage as follows:
\begin{align}
    %R_c^*, \bm{t}_c^* = \argmax_{R_c, \bm{t}_c} \sum_{\subalign{0 &\leq x_i \leq W \\ \land 0 &\leq y_i \leq H \\ \land \lnot &\text{Occ}_i}} \sigma_{\text{mesh}, i}
    R_c^*, \bm{t}_c^* = \argmax_{R_c, \bm{t}_c} \sum_{\bm{v}_i \in V_\text{ML}} \sigma^\text{mesh}_i f^\text{visible}_i \label{eq:camera_view_evaluation}
\end{align}
where, $f^\text{visible}_i$ is the boolean for visibility of the 3D position of $i$-th vertex $\bm{v}_i$ from the mesh vertices $V$.
$f^\text{visible}_i$ is calculated by the camera projection function $\pi$ that projects a 3D position to the 2D image coordinates and checks self-occlusion as follows:
\begin{align}
    f^\text{visible}_i &= \begin{cases}
    1 & \text{if} \ 0 \leq x_i \leq W, 0 \leq y_i \leq H, \text{Occ}_i = 0\\ 
    0 & \text{otherwise}
    \end{cases}\\
    &x_i, y_i, \text{Occ}_i = \pi(\bm{v}_i; V, R_c, \bm{t}_c),
\end{align}
where $x_i, y_i$ are projected positions on the 2D image coordinates. $W, H$ are the width and height of the input camera image. $\text{Occ}_i$ denotes the boolean for self-occlusion as $\text{Occ}_i \in \{0, 1\}$ \added[id=RevAB, comment={A.3, A.6, B.4}]{calculated based on the Z-buffer method. Specifically, we calculate the distances to all faces on each camera ray to determine whether faces are visible. $Occ_i$ on each point is 0 if the point is contained by visible faces or 1 if contained by occluded faces.} Although Eq.~(\ref{eq:camera_view_evaluation}) can evaluate camera viewpoints, this is not differentiable and unsuitable for backpropagation. Therefore, the camera viewpoint is optimized by picking the best from a predefined set of viewpoints. 
The method is explained with one camera for brevity, but one can easily extend the viewpoint optimization to multiple cameras.

\noindent\textbf{Touch and ranging sensor placement optimization:} For touch and ranging sensors that measure the body surface, its placements $(R_1, \bm{t}_1), \dots, (R_K, \bm{t}_K)$ are optimized to complement the limited information from the optimized viewpoint.
Specifically, we optimize the placements to measure the most uncertain vertex within kinematically possible placements by solving inverse kinematics (IK) as follows:
\begin{align}
    q_i, \text{Suc}_i =& \text{IK}\big(\bm{v}_{i} ; (R_c^*, \bm{t}_c^*), (R_1^*, \bm{t}_1^*), \dots, (R_{k-1}^*, \bm{t}_{k-1}^*)\big) \\
     i^* = &\argmax_{i \in [1, 6890]} \sigma^\text{mesh}_i \text{Suc}_i \\
     R_k^*, \bm{t}_k^* =& \text{FK}(q_{i^*})
\end{align}
where IK is the inverse kinematics function that takes a 3D position $\bm{v}_i$ of $i$-th mesh vertex and already optimized sensor placements $(R_c^*, \bm{t}_c^*), (R_1^*, \bm{t}_1^*), \dots, (R_{k-1}^*, \bm{t}_{k-1}^*)$ as input, and outputs the joint parameters $q_i$ which sensing the 3D position $\bm{v}_i$ of $i$-th mesh vertex with the target sensor and $\text{Suc}_i \in {0, 1}$ denotes the boolean for whether there is feasible joint parameters $q_i$ satisfying the all optimized sensor placements. FK is the forward kinematics function that calculates the target sensor position $R_k, \bm{t}_k$ from the joint parameters $q$.
%If the target sensor has limited freedom for placement due to the robot's task such as supporting the human body, we simply perform the optimization within the allowed joint range.
%We can easily extend the active measurement framework above to sense the multiple mesh vertices using the pressure sensor array.
After this optimization, we fuse the multimodal cues to the estimated pose.

\subsection{Sensor Fusion}
\label{subsec:fusion}
Our sensor fusion improves estimated poses by aligning the estimated mesh $V$ consistent with the \replaced[id=RevB, comment={B.5}]{depth measurements of human body surface}{body surface measurements} from currently optimized $k$ sensors $\bm{m}_1, \dots, \bm{m}_k$.
The proposed sensor fusion has two stages: global position offset and local pose optimization.

\noindent\textbf{Global position offset:} We offset the global position $\bm{t}_p$ because camera-based estimated poses contain depth ambiguity due to an unknown person's height and inaccurate camera intrinsics.
The offset is the mean vector between the measurements $\bm{m}_1, \dots, \bm{m}_k$ to the nearest mesh vertex $\bm{w}_1, \dots, \bm{w}_k$ from $V_\text{ML}$ as follows:
\begin{align}
    \bm{t}_\text{offset} &= \mathbb{E}_{(\bm{\theta}, \bm{\beta}) \sim p(\bm{\theta}, \bm{\beta}|I)} \bigg[\cfrac{1}{k} \sum_{k'=1}^k \bm{w}_{k'} - \bm{m}_{k'}\bigg] \\
    \bm{w}_k &= \argmin_{\bm{v}_i} || \bm{v}_i - \bm{m}_k ||
\end{align}

\noindent\textbf{Local pose optimization:} After calculating offset, we optimize each 
 sampled pose parameter $\bm{\theta}$ by minimizing the distance between the body surface and the sensor measurements via backpropagation as follows:
\begin{align}
    \bm{\theta}^* &= \argmin_{\bm{\theta}} L \\
    L &= -p(\bm{\theta}, \bm{\beta}|I) + \cfrac{\alpha}{k} \sum_{k'=1}^k \bm{w}_{k'} - \bm{m}_{k'} - \bm{t}_\text{offset}
\end{align}
where $\alpha$ is a weighting parameter set as $0.01$.
Using each optimized pose parameter $\bm{\theta}^*$, we update the positional variance $\bm{\sigma}^\text{mesh}$ as described in Sec.~\ref{subsec:uncertainty} and optimize the next sensor placement $R_{k+1}, \bm{t}_{k+1}$ by repeating the process iteratively.
The updated positional variance $\bm{\sigma}^\text{mesh}$ is expected to be small around the sensor measurements thanks to optimization.
Therefore, the next measured point is a different uncertain vertex based on $\bm{\sigma}^\text{mesh}$.
The maximum-likelihood pose with sensor fusion $\bm{\theta}^*_\text{ML}$ is used for evaluation.

\section{EXPERIMENTS}
We benchmarked the proposed method on the pose estimation dataset with occlusion by simulated active measurements in Sec.~\ref{subsec:benchmark}.
For real-world experiments, settings are described in Sec.~\ref{subsec:real_world_setup}.
We report the quantitative results of the real-world experiments in Sec.~\ref{subsec:real_world_results}.
The qualitative results under more constraints are described in Sec.~\ref{subsec:real_world_scenarios}.\footnote{\added[id=RevA,comment={A.9}]{The code is publicly available at \url{https://github.com/meaten/HMRinCloseProximity}.}
}

\subsection{Quantitative Comparison on Benchmark}
\label{subsec:benchmark}
We compare the proposed and previous methods on 3DPW-OC~\cite{DBLP:conf/eccv/MarcardHBRP18, zhang2020object}, the standard benchmark for human pose estimation with occlusion.
3DPW-OC contains pairs of images and ground truth SMPL mesh annotations.
We omitted the camera viewpoint optimization because we cannot optimize and retake the input images in the benchmark dataset.
For touch and ranging sensors, we simulated the measured vertices by sampling from ground truth mesh \added[id=RevB, comment={B.6}]{and adding a artificial Gaussian noise $\sigma=0.02$[m]. The $\sigma$ value is determined based on the 95 percent confidence interval $2\sigma=0.04$ [m] of the range sensor accuracy within $\pm0.04$[m] and HSR's hand accuracy within $\pm0.02$[m].}
% based on the positional variance $\bm{\sigma}^\text{mesh}$ as described in Sec.~\ref{subsec:active_measurement}.

\begin{table}[t]
\centering
\caption{{\bf Quantitative results on 3DPW-OC in [mm].} \added[id=RevA, comment={A.1}]{RM and AM denote random measurement and our active measurement. CS and SF denote the closest sample selection and our sensor fusion.}}
\begin{tabular}{ll|rr}
Method & Measured Vertices & MPJPE$\downarrow$ & PA-MPJPE$\downarrow$ \\ \hline
I2L-MeshNet~\cite{DBLP:conf/eccv/MoonL20} & n=0 & 92.0 & 61.4 \\
SPIN~\cite{DBLP:journals/tog/HuangKABHP18} & n=0 & 95.5 & 60.7 \\
PyMAF~\cite{DBLP:conf/iccv/ZhangTZOL0S21} & n=0 & 89.6 & 59.1 \\
ROMP~\cite{DBLP:conf/iccv/SunBLFB021} & n=0 & 91.0 & 62.0 \\
OCHMR~\cite{DBLP:conf/cvpr/KhirodkarTK22} & n=0 & 112.2 & 75.2 \\
PARE~\cite{DBLP:conf/iccv/KocabasHHB21} & n=0 & 83.5 & 57.0 \\
3DCrowdNet~\cite{DBLP:conf/cvpr/ChoiMPL22} & n=0 & 83.5 & 57.1 \\
JOTR~\cite{DBLP:conf/iccv/LiYWMZY23} & n=0 & 75.7 & 52.2 \\
ProHMR~\cite{DBLP:conf/iccv/KolotourosPJD21} & n=0 & 92.9 & 62.2 \\ \hline
\multirow{4}{*}{ProHMR + RM + CS} & n=3 & 89.0 & 62.3 \\
 & n=5 & 87.1 & 60.7 \\
 & n=10 & 84.9 & 58.8 \\
 & n=30 & 82.7 & 57.1 \\ \hline
\multirow{4}{*}{ProHMR + RM + SF} & n=3 & 86.0 & 61.6 \\
 & n=5 & 81.6 & 58.4 \\
 & n=10 & 75.6 & 53.7 \\
 & n=30 & 68.5 & 47.6 \\ \hline
\multirow{4}{*}{ProHMR + AM + CS} & n=3 & 88.3 & 61.1 \\
 & n=5 & 86.3 & 59.5 \\
 & n=10 & 84.1 & 57.9 \\
 & n=30 & 82.7 & 57.2 \\ \hline
\multirow{4}{*}{ProHMR + AM + SF} & n=3 & 83.5 & 59.2 \\
 & n=5 & 78.0 & 55.1 \\
 & n=10 & 71.1 & 49.3 \\
 & n=30 & \textbf{67.2} & \textbf{46.1}
\end{tabular}
\label{tab:3dpw-oc}
\end{table}

The second column of the Tab.~\ref{tab:3dpw-oc} denotes the number of measured vertices.
We used ProHMR~\cite{DBLP:conf/iccv/KolotourosPJD21} as the off-the-shelf probabilistic human mesh recovery.
However, model weights were trained from scratch with a focal length of $2200$ to deal with the changing focal length of input images~\cite{DBLP:conf/eccv/KissosFGMOK20}. 
\added[id=RevA, comment={A.1}]{To validate the proposed active measurement and sensor fusion separately, we compare these with naive measurements and fusion methods. Therefore, we replace the proposed method with random measurement (RM) and closest sample selection (CS). RM randomly measures the mesh points, unlike our active measurement (AM) choosing the most uncertain mesh points. CS selects the pose closest to the measurements within samples from the pose distribution $p(\bm{\theta}, \bm{\beta}|I)$, unlike our sensor fusion (SF) aligning the closest pose via optimization.}
We report the mean per joint position error (MPJPE) in Tab.~\ref{tab:3dpw-oc}.
\added[id=RevA, comment={A.1, A.4, A.7}]{The results show the methods with our sensor fusion significantly increase the accuracy with sparse but reliable sensor measurements.
Our active measurement can further improve the accuracy of our sensor fusion.}
\added[id=RevA, comment={A.8}]{The computational costs of our active measurement and sensor fusion are $6.4\pm 2.0$[ms] and $2417.7\pm 814.4$[ms] with Intel Xeon Gold CPU and one NVIDIA RTX TITAN GPU.}

\begin{figure}[t]
\centering
\includegraphics[width=0.90\linewidth]{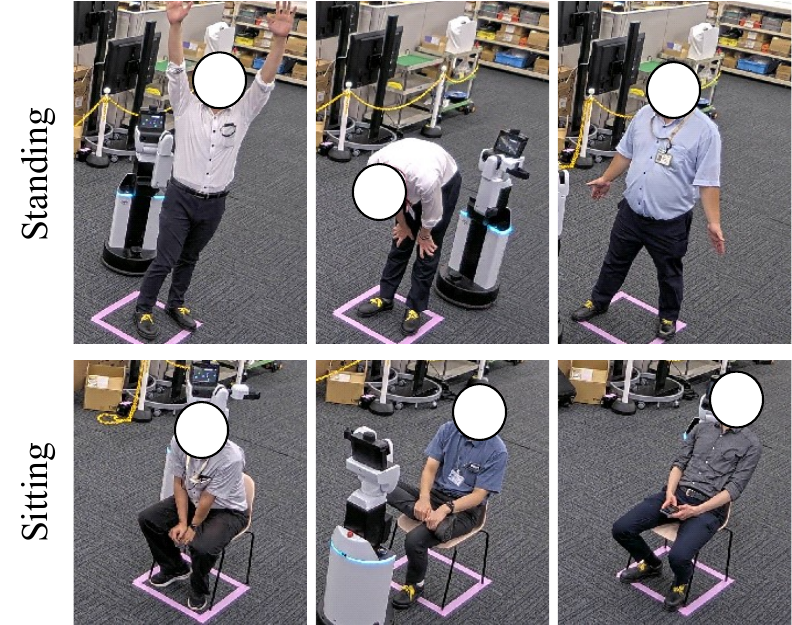}
\caption{\textbf{Evaluated poses in two pose groups: standing and sitting.}}
\label{fig:poses}
\end{figure}

\subsection{Experimental Setup of Real-World Experiments}
\label{subsec:real_world_setup}
%Real-world quantitative experiments were conducted to show the effectiveness of our proposed method.
%We also conducted qualitative experiments on practical scenarios of a person occluded by a blanket and standing aid with the robot arm.
Toyota Human Support Robot (HSR)~\cite{DBLP:conf/siggraph/YamamotoNKOI18, yamamoto2019development}, shown in the bottom of Fig.~\ref{fig:teaser}, was used in the real-world experiments.
The HSR has RGB cameras, a force sensor in the wrist joint, and 2D LiDAR.
The RGB camera has 5 degrees of freedom by the omnidirectional moving base, the lifting joint, and the pan/tilt joint.
The force sensor has 3 degrees of freedom by the yaw rotation of the omnidirectional moving base and the robot arm.
The 2D LiDAR has 3 degrees of freedom by the omnidirectional moving base.
We applied a leg detection filter to the lower-equipped 2D LiDAR measurements of HSR to exclude the walls and other objects.
We used the same model weights of ProHMR as Sec.~\ref{subsec:benchmark}.
For the quantitative evaluation, the ground truth poses are measured by markerless human motion capture EasyMocap~\cite{easymocap}.
EasyMocap utilizes multiple calibrated RGB environmental cameras to satisfy enough accuracy to create the human motion dataset ZJU-Mocap~\cite{DBLP:conf/cvpr/FangSDBZ21}.
We used six calibrated cameras to utilize EasyMocap.

\subsection{Quantitative Results on Real-World Experiments}
\label{subsec:real_world_results}
\begin{figure}[t]
\centering
\includegraphics[width=0.95\linewidth]{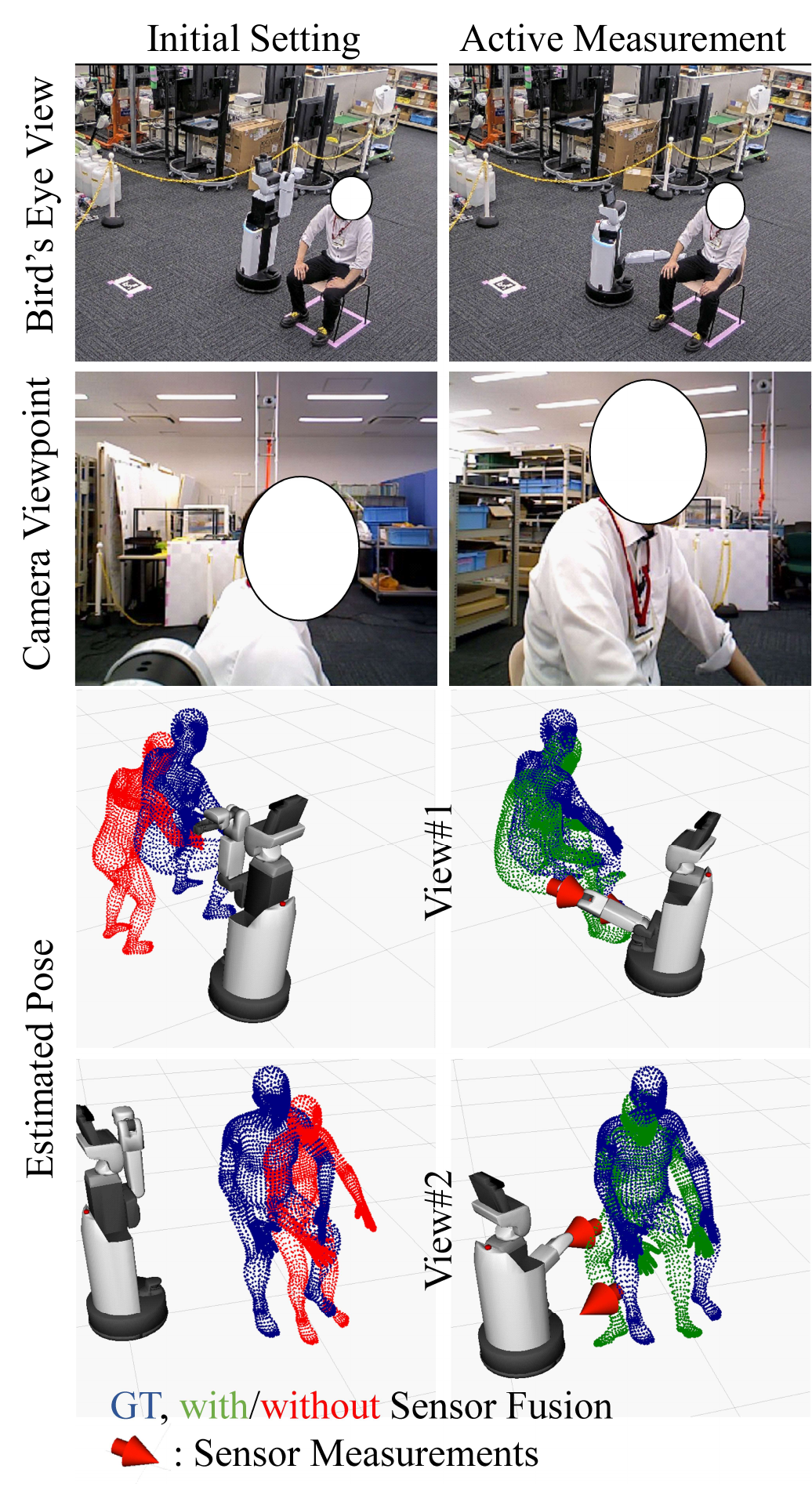}
\caption{\textbf{Visualization of quantitative evaluation.}}
\label{fig:quantitative}
\end{figure}

\begin{figure}[t]
\centering
\includegraphics[width=0.95\linewidth]{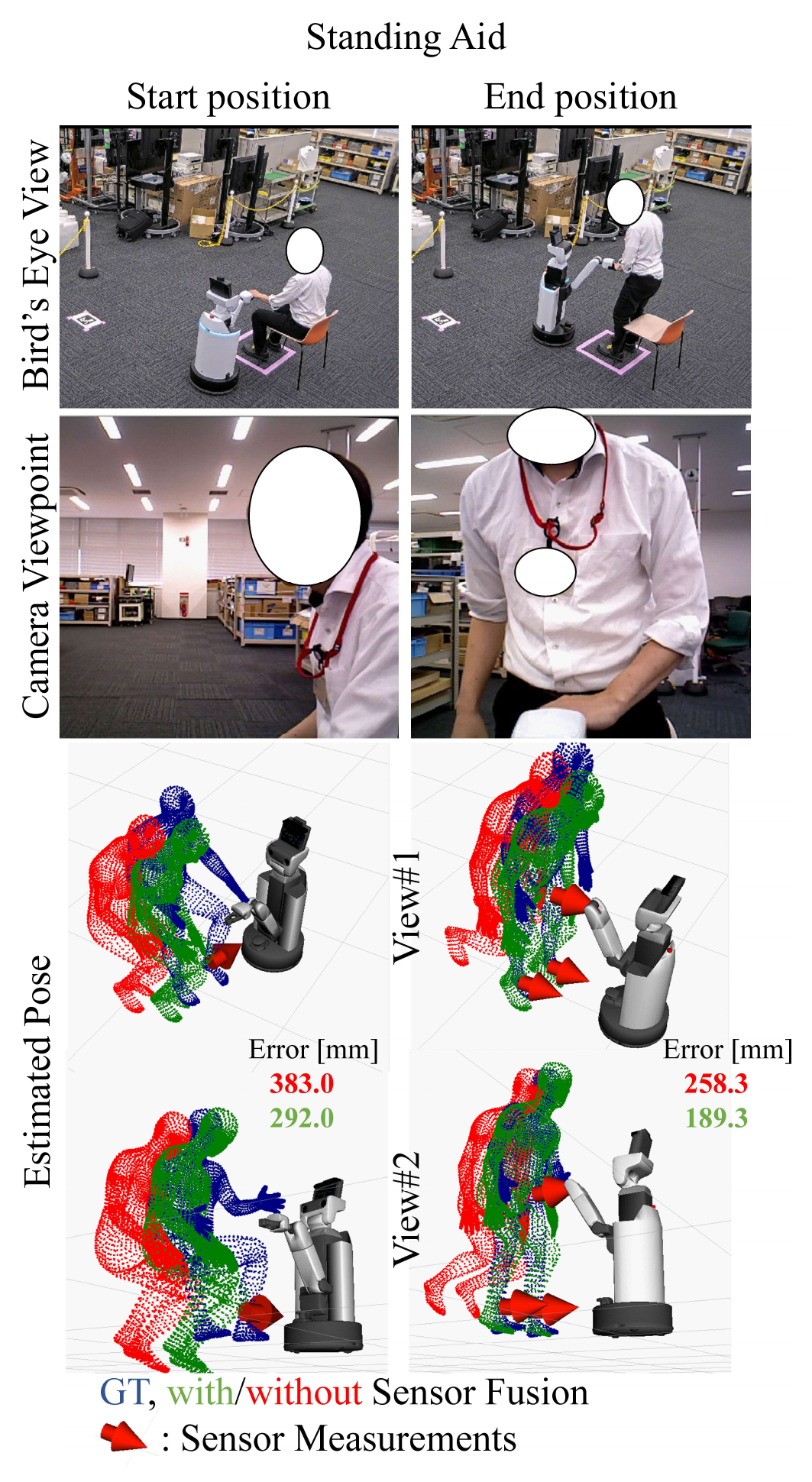}
\caption{\textbf{Visualization of the standing aid scenario.}}
\label{fig:standing}
\end{figure}

We evaluated the accuracy of the estimated poses of ten subjects. Eight poses in each of the two pose groups were captured for each subject.
Therefore, $160$ poses are evaluated in total.
HSR estimates the pose with our active measurement and sensor fusion from a constant 0.7-meter distance to the subject, where 3D point cloud sensor measurements frequently drop due to the out-of-range depending on the posture.
\added[id=RevB, comment={B.9}]{HSR starts from 8 different positions surrounding the subject with 45 degree intervals.}
The viewpoint is optimized to be one of eight directions with 45-degree intervals and one of three heights by grid search as described in Sec.~\ref{subsec:active_measurement}.
The subjects are directed to take poses in two groups: standing and sitting, as shown in Fig.~\ref{fig:poses}.
The subjects are 163-187 centimeter-tall adults with around 60-110 kilogram weights.
The project of this experiment obtained approval from the Research Ethics Review Committee of Toyota Motor Corporation on July 31st, 2023.

We report the mean Euclidean distance error in Tab.~\ref{tab:quantitative_standing} and Tab.~\ref{tab:quantitative_sitting}.
Unlike previous standard benchmarks such as 3DPW-OC, as shown in Tab.~\ref{tab:3dpw-oc}, we didn't offset the global translation error for a more realistic assumption.
JOTR~\cite{DBLP:conf/iccv/LiYWMZY23}, which achieved the best performance among the camera-based methods on 3DPW-OC, results in high Euclidean distance error.
This poor performance is because severe truncations as shown in the second row of Fig.~\ref{fig:poses} cause poorly estimated 2D poses that JOTR heavily depends on during inference.
On the other hand, the proposed method achieved the lowest Euclidean distance error averaged on all joints in both pose groups.
Our camera active measurement can significantly decrease the estimation error.
Sensor fusion of touch sensors can reduce the error of all joints by sensing the whole body relatively freely and complementing limited information from the camera.
In contrast, 2D LiDAR mainly improves the estimation of leg poses by sensing the occupation near the floor.
2D LiDAR can reduce the error of ankle joints to half in the standing pose group.
The estimated poses are visualized in Fig.~\ref{fig:quantitative}.
%The estimated poses are improved by covering more body surface within the camera image and touching the thigh with the robot arm.

\begin{table*}[t]
\caption{\textbf{Quantitative results on standing poses.}}
\centering
\begin{tabular}{lccc|rrrrrrr}
\multirow{2}{*}{Base Model} & \multicolumn{3}{c|}{Active Measurement \& Sensor Fusion} & \multicolumn{7}{c}{Euclidean Distance Error [mm]} \\ \cline{2-11} 
 & Camera & Touch Sensor & 2D LiDAR & \multicolumn{1}{c}{Head} & \multicolumn{1}{c}{Pelvis} & \multicolumn{1}{c}{R Wrist} & \multicolumn{1}{c}{L Wrist} & \multicolumn{1}{c}{R Ankle} & \multicolumn{1}{c|}{L Ankle} & \multicolumn{1}{c}{All Joints} \\ \hline
JOTR~\cite{DBLP:conf/iccv/LiYWMZY23} & \multicolumn{1}{l}{} & \multicolumn{1}{l}{} & \multicolumn{1}{l|}{} & 579.1 & 513.5 & 532.3 & 537.3 & 540.0 & \multicolumn{1}{r|}{548.7} & 5298 \\ \hline
\multirow{5}{*}{ProHMR~\cite{DBLP:conf/iccv/KolotourosPJD21}} & \multicolumn{1}{l}{} & \multicolumn{1}{l}{} & \multicolumn{1}{l|}{} & 400.4 & 370.7 & 437.3 & 409.1 & 533.4 & \multicolumn{1}{r|}{547.5} & 418.1 \\
 & \checkmark &  &  & 210.6 & 224.4 & 274.0 & 275.4 & 428.7 & \multicolumn{1}{r|}{442.5} & 266.5 \\
 & \checkmark & \checkmark &  & \textbf{210.2} & \textbf{201.5} & \textbf{256.6} & \textbf{264.3} & 387.1 & \multicolumn{1}{r|}{400.8} & 249.8 \\
 & \checkmark &  & \checkmark & 276.5 & 215.9 & 288.2 & 298.8 & 210.7 & \multicolumn{1}{r|}{\textbf{210.5}} & 243.0 \\
 & \checkmark & \checkmark & \checkmark & 269.1 & 203.9 & 275.8 & 287.6 & \textbf{202.8} & \multicolumn{1}{r|}{215.8} & \textbf{235.9}
\end{tabular}
\label{tab:quantitative_standing}
\end{table*}

\begin{table*}[t]
\caption{\textbf{Quantitative results on sitting poses.}}
\centering
\begin{tabular}{lccc|rrrrrrr}
\multirow{2}{*}{Base Model} & \multicolumn{3}{c|}{Active Measurement \& Sensor Fusion} & \multicolumn{7}{c}{Euclidean Distance Error [mm]} \\ \cline{2-11} 
 & Camera & Touch Sensor & 2D LiDAR & \multicolumn{1}{c}{Head} & \multicolumn{1}{c}{Pelvis} & \multicolumn{1}{c}{R Wrist} & \multicolumn{1}{c}{L Wrist} & \multicolumn{1}{c}{R Ankle} & \multicolumn{1}{c|}{L Ankle} & \multicolumn{1}{c}{All Joints} \\ \hline
JOTR~\cite{DBLP:conf/iccv/LiYWMZY23} & \multicolumn{1}{l}{} & \multicolumn{1}{l}{} & \multicolumn{1}{l|}{} & 543.3 & 582.3 & 528.0 & 586.6 & 795.3 & \multicolumn{1}{r|}{819.9} & 584.7 \\ \hline
\multirow{5}{*}{ProHMR~\cite{DBLP:conf/iccv/KolotourosPJD21}} & \multicolumn{1}{l}{} & \multicolumn{1}{l}{} & \multicolumn{1}{l|}{} & 452.7 & 514.0 & 515.2 & 510.6 & 791.3 & \multicolumn{1}{r|}{759.5} & 572.4 \\
 & \checkmark &  &  & 206.3 & 273.9 & 322.1 & 298.5 & 539.1 & \multicolumn{1}{r|}{534.4} & 322.9 \\
 & \checkmark & \checkmark &  & 219.3 & 212.6 & 293.0 & 264.1 & 446.9 & \multicolumn{1}{r|}{449.9} & 284.2 \\
 & \checkmark &  & \checkmark & \textbf{206.1} & 199.1 & 267.6 & 254.2 & 442.9 & \multicolumn{1}{r|}{415.4} & 274.9 \\
 & \checkmark & \checkmark & \checkmark & 210.3 & \textbf{192.7} & \textbf{267.1} & \textbf{249.5} & \textbf{419.4} & \multicolumn{1}{r|}{\textbf{405.8}} & \textbf{270.0}
\end{tabular}
\label{tab:quantitative_sitting}
\end{table*}

\subsection{Estimation Accuracy under Practical Constraints}
\label{subsec:real_world_scenarios}
The quantitative evaluations prove the effectiveness of our method with no restriction on camera and sensor placements in Sec.~\ref{subsec:real_world_results}.
However, the practical pHRI scenarios have several restrictions depending on the target task.

\begin{figure*}[t]
\centering
\includegraphics[width=\linewidth]{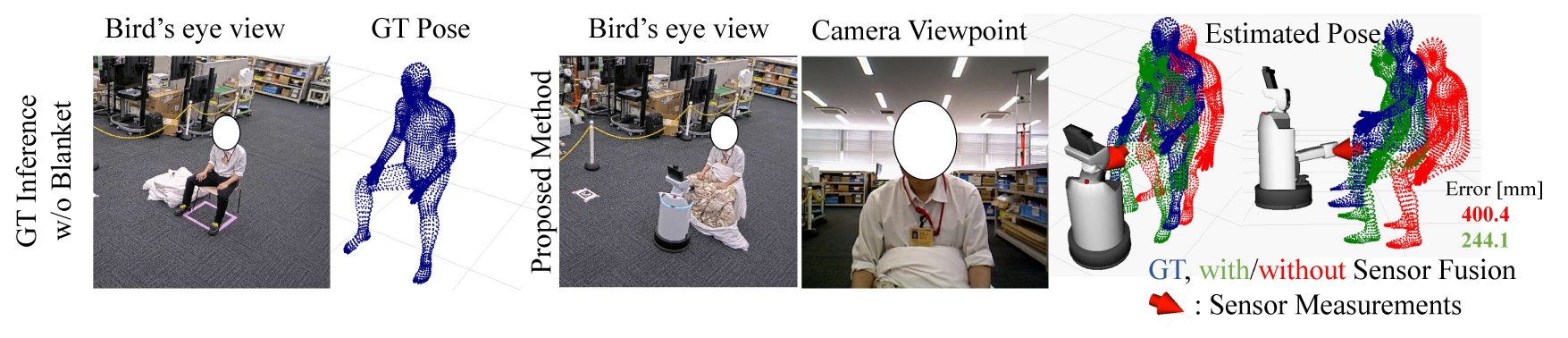}
\caption{\textbf{Visualization of the scenario of the target person with a blanket.}}
\label{fig:blanket}
\end{figure*}

\noindent \textbf{Standing Aid Scenario.} 
When assistive robots support human bodies or hand something to a person, robot arms must reach certain positions \added[id=RevB, comment={B.10}]{while estimating accurate poses for fall detection}.
Therefore, the placement of the touch sensor equipped at the end effector is constrained in standing aid scenarios, as shown in the top row of Fig.~\ref{fig:standing}.
We mocked this situation and experimented with the standing aid scenario.
The HSR optimizes the camera viewpoint while its end effector follows a specific lifting trajectory.
The estimated poses are visualized at the bottom of Fig.~\ref{fig:standing}.
The proposed method estimates the poses closer to the ground truth even with the placement constraint on the touch sensor.
The viewpoint is improved from the side low view to the front high view to cover the larger body while following the target person standing up.
The sensor fusion of 2D LiDAR also improves the leg positions.

\noindent \textbf{Occluded by a Blanket Scenario.}
Active measurement on camera viewpoints may not be effective in several cases.
For example, the target person is sitting under a comforter on a bed or blanket. \added[id=RevB, comment={B.10}]{Assistive robots must estimate accurate poses for posture evaluation to prevent bedsores.}
Furthermore, wearing loose-fitting clothes such as dresses, hospital patient gowns, and kimonos also affects viewpoint optimization.
However, the proposed method can still estimate the accurate pose thanks to multimodality because touch sensors are robust to these occlusions and reliably obtain the body surface information.

We experimented with a scenario where a blanket occluded the target person.
HSR estimates the poses of the target person with a blanket on the person's lower body.
%The touch sensor placement is optimized while the camera viewpoint is fixed.
The 2D LiDAR measurements were not utilized because the leg detection filter could not detect the leg due to the broad blanket surface.
\added[id=RevB, comment={B.11}]{All leg detections around the subjects are filtered out by confidence threshold due to too large diameters of blanket surface for human legs.}
The estimated poses are visualized in Fig.~\ref{fig:blanket}.
The proposed method estimates the poses closer to the ground truth by utilizing touch sensor measurement despite poor information from the camera.

\section{DISCUSSION}
\label{sec:discussion}
In this section, we discuss the validity of our experimental setting with a single camera and additional sensors.

\noindent \textbf{Multiple Calibrated Environmental cameras} may estimate accurate poses as EasyMocap~\cite{easymocap} and ZJU-Mocap dataset~\cite{DBLP:conf/cvpr/FangSDBZ21}.
However, environmental cameras face serious privacy problems.
Usually, no environmental camera is allowed in restrooms because of privacy violations to others despite the need for standing aid of elderly or physically challenged people.
Furthermore, environmental cameras are not available outside the buildings.
Utilizing only a camera mounted to the standalone robot body is straightforward.

\noindent \textbf{Multiple cameras on multiple robot arms} may also estimate accurate poses by triangulation and more visible area.
However, due to cost constraints, robots usually don't equip high DoF arms only for camera observation as Spot of Boston Dynamics, mobile manipulator of Everyday Robots, and Mobile ALOHA~\cite{fu2024mobile}.
%However, multiple cameras on multiple robot arms face complicated kinematic constraints.
%In physical human-robot interaction scenarios, there are already several robot arms used for interactions such as handing objects and supporting a human body.
%Thus, it is difficult to move multiple cameras with arms to informative viewpoints under the constraints of a stable center of mass and free from self-collisions.
Although we didn't conduct experiments with multiple cameras for the above reason, our method can easily be extended to accept multiple images.

\section{CONCLUSION}
In this paper, we proposed an active measurement and sensor fusion framework for human pose estimation that can estimate accurate human poses in close proximity.
Experimental results demonstrated that our method achieved higher accuracy than methods without active measurement or sensor fusion, even with severely truncated camera images.

Our proposed method has several limitations: slow inference and not utilizing temporal information.
Our method takes $0.5$ seconds to estimate poses due to the optimization in the sensor fusion.
We can achieve faster inference by replacing the optimization with regression~\cite{oba2023data}.
In addition to this, we can incorporate temporal information or a previously estimated pose to skip the several processes in our proposed method.
\deleted[id=RevB, comment={B.13}]{Our future works are achieving higher accuracy by using other sensors such as a pressure sensor array, proximity sensors, and thermometers. }

%\section*{APPENDIX}
%\section*{ACKNOWLEDGMENT}

%%%%%%%%%%%%%%%%%%%%%%%%%%%%%%%
% CAUTION! Need to check the bib style e.g. font size because the bib style is added
%%%%%%%%%%%%%%%%%%%%%%%%%%%%%%%
\bibliographystyle{ieee_fullname}
\bibliography{root}

\end{document}